\documentclass{egpubl}
\usepackage{pg2025s}

\WsConferencePaper

\usepackage[T1]{fontenc}
\usepackage{dfadobe}  

\usepackage{cite}  
\BibtexOrBiblatex
\electronicVersion
\PrintedOrElectronic
\ifpdf \usepackage[pdftex]{graphicx} \pdfcompresslevel=9
\else \usepackage[dvips]{graphicx} \fi

\usepackage{egweblnk} 

\title[CMI-MTL: Cross-Mamba interaction based multi-task learning for medical visual question answering]%
      {CMI-MTL: Cross-Mamba interaction based multi-task learning for medical visual question answering}

\author[Jin et al.]
{\parbox{\textwidth}{\centering 
Qiangguo Jin$^{1,2}$\orcid{0000-0002-1781-1067}, 
Xianyao Zheng$^{1}$, 
Hui Cui$^{3}$\orcid{0000-0001-8224-4698}, 
Changming Sun$^{4}$\orcid{0000-0001-5943-1989}, 
Yuqi Fang$^{5}$\orcid{0000-0002-8769-496X}, 
Cong Cong$^{6}$\orcid{0000-0002-8192-6731}, 
Ran Su$^{7}$, 
Leyi Wei$^{8}$\orcid{0000-0003-1444-190X},\\ 
Ping Xuan$^{9}$, 
and Junbo Wang$^{1}$\thanks{Corresponding author}
        }
        \\
{\parbox{\textwidth}{
\centering 
    $^1$School of Software, Northwestern Polytechnical University, Shaanxi, China\\
    $^2$Yangtze River Delta Research Institute of Northwestern Polytechnical University, Taicang, China\\
    $^3$Department of Computer Science and Information Technology, La Trobe University, Melbourne, Australia\\
    $^4$CSIRO Data61, Sydney, Australia\\
    $^5$School of Intelligence Science and Technology, Nanjing University, Suzhou, China\\
    $^6$Australian Institute of Health Innovation (AIHI), Macquarie University, Australia\\
    $^7$School of Computer Software, College of Intelligence and Computing, Tianjin University, Tianjin, China\\
    $^8$Centre for Artificial Intelligence driven Drug Discovery, Faculty of Applied Science, Macao Polytechnic University, Macao Special Administrative Region of China\\
    $^{9}$Department of Computer Science, School of Engineering, Shantou University, Guangdong, China
       }
}
}

\usepackage{amsmath}
\usepackage{amsbsy}
\usepackage{multirow} 
\usepackage{amssymb}
\usepackage{hyperref}
\usepackage{marvosym}
\begin{document}


\maketitle
\begin{abstract}
Medical visual question answering (Med-VQA) is a crucial multimodal task in clinical decision support and telemedicine.
Recent self-attention based methods struggle to effectively handle cross-modal semantic alignments between vision and language. Moreover, classification-based methods rely on predefined answer sets. Treating this task as a simple classification problem may make it unable to adapt to the diversity of free-form answers and overlook the detailed semantic information of free-form answers. In order to tackle these challenges, we introduce a \textbf{C}ross-\textbf{M}amba \textbf{I}nteraction based \textbf{M}ulti-\textbf{T}ask \textbf{L}earning (CMI-MTL) framework that learns cross-modal feature representations from images and texts. CMI-MTL comprises three key modules: fine-grained visual-text feature alignment (FVTA), cross-modal interleaved feature representation (CIFR), and free-form answer-enhanced multi-task learning (FFAE). FVTA extracts the most relevant regions in image-text pairs through fine-grained visual-text feature alignment. CIFR captures cross-modal sequential interactions via cross-modal interleaved feature representation. FFAE leverages auxiliary knowledge from open-ended questions through free-form answer-enhanced multi-task learning, improving the model's capability for open-ended Med-VQA.
Experimental results show that CMI-MTL outperforms the existing state-of-the-art methods on three Med-VQA datasets: VQA-RAD, SLAKE, and OVQA. Furthermore, we conduct more interpretability experiments to prove the effectiveness. The code is publicly available at \url{https://github.com/BioMedIA-repo/CMI-MTL}.

\ccsdesc[300]{Computing methodologies~Medical visual question answering}
\ccsdesc[300]{Computing methodologies~Multi-task learning}
\ccsdesc[100]{Computing methodologies~Cross-Mamba interaction}

\printccsdesc   
\end{abstract}  

\section{Introduction}
Medical visual question answering (Med-VQA) focuses on deriving correct answers from given medical images and the associated professional questions, serving a critical role in clinical decision support and telemedicine~\cite{lin2023medical}.
Unlike general visual question answering, Med-VQA primarily relies on data from medical imaging modalities such as X-rays, CT, MRI, and ultrasound~\cite{ben2019vqa}. Moreover, the questions focus on medical diagnosis, treatment, and anatomical structure recognition, requiring a deep understanding of medical terminology, imaging modalities, and disease patterns~\cite{seenivasan2022surgical}. However, mapping heterogeneous data into a shared representation space for holistic understanding remains challenging, as interpreting medical images demands specialized knowledge to discern subtle yet clinically significant details.

\begin{figure*}[htb]
    \centering
    \includegraphics[width=1\linewidth]{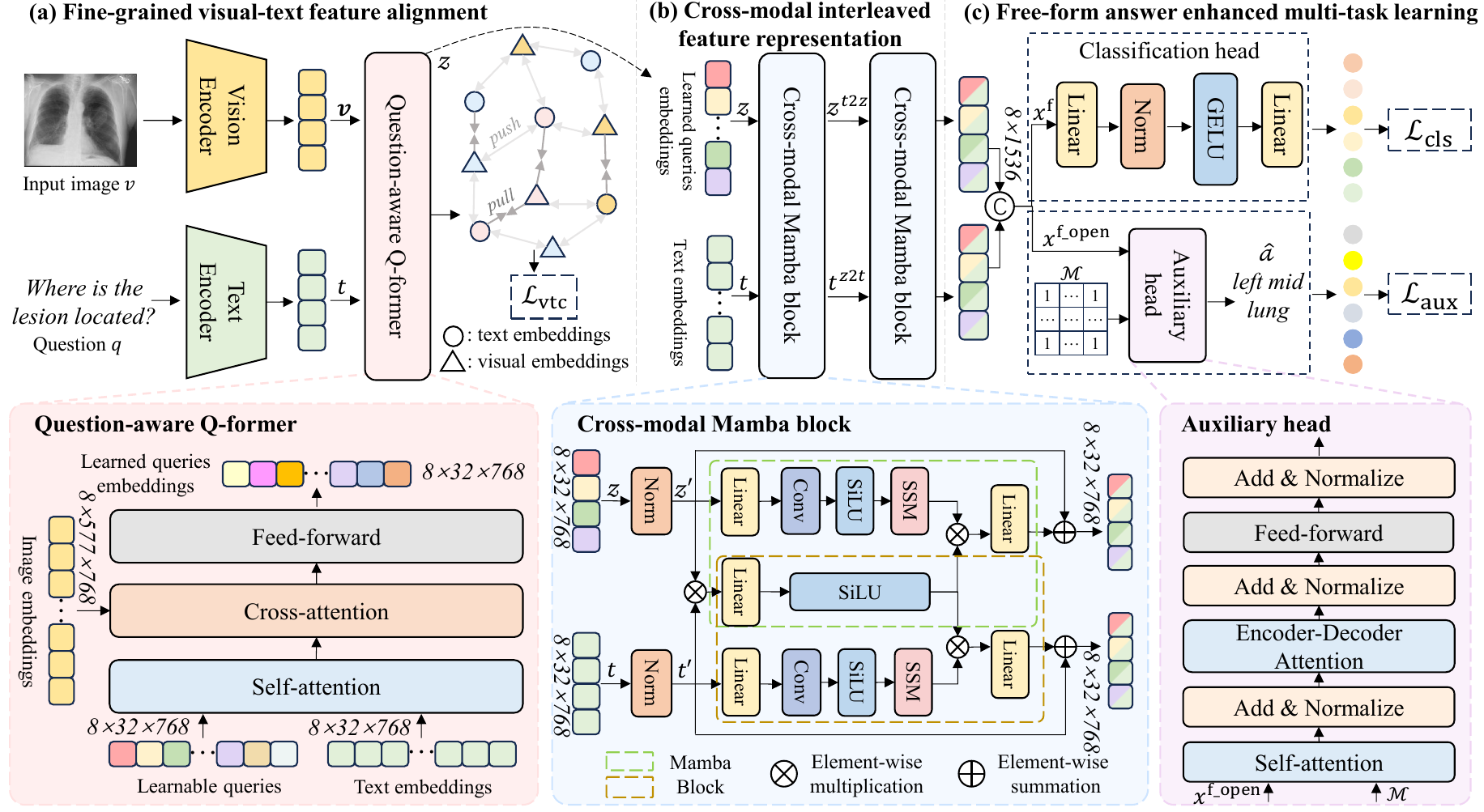}
    \caption{The overall architecture of CMI-MTL. The CMI-MTL consists of (a) Fine-grained visual-text feature alignment, (b) Cross-modal interleaved feature representation, and (c) Free-form answer enhanced multi-task learning.}
    \label{fig:architecture}
\end{figure*}

To handle the medical data understanding challenge, recent methods~\cite{nguyen2019overcoming,do2021multiple,li2023self,ossowski2023retrieving,gu2024lapa} have been proposed, leveraging visual-language interactions for learning. For example, Do et al.~\cite{do2021multiple} proposed a multiple meta-model quantification method that effectively learns meta-annotations for the Med-VQA task. Gu~et~al.~\cite{gu2024lapa} leveraged latent prompts to enhance understanding within a Transformer-based~\cite{vaswani2017attention} framework. Li~et~al.~\cite{li2023self} applied masked image modeling, masked language modeling, image-text matching, and image-text alignment to enhance model generalization.

However, current Med-VQA methods face several limitations. First, merely training Transformer-based large-scale language models on small Med-VQA datasets can lead to overfitting~\cite{PAN2022109763}. Second, while transformer-based methods effectively capture global relationships via self-attention, they may struggle with the vagueness of cross-modal information interactions and lack the necessary inductive biases for sequential data, such as long questions (e.g., \textit{Is there left distal radius fracture or left ulnar styloid process avulsion fracture in this image?})~\cite{10256025}. Finally, most methods treat both closed-ended (e.g., Q: \textit{Are regions of the brain infarcted?} $\to$ A: \textit{Yes}) and open-ended (e.g., Q: \textit{In which two ventricles can calcifications be seen on this CT scan?} $\to$ A: \textit{The 3rd ventricle and the lateral ventricles}) questions equally as classification tasks, mapping them to a predefined answer sets~\cite{9434063}. While these models achieve good performance, they are limited in fully leveraging the semantic information in open-ended questions. When handling free-form answers in datasets VQA-RAD~\cite{lau2018dataset} and SLAKE~\cite{liu2021slake}, they are unable to flexibly accommodate the diversity and variability of these answers. This lack of adaptability restricts the ability of the model to accurately represent and utilize the rich semantic content within the answers.

To address these challenges, we propose a Cross-Mamba Interaction based Multi-Task Learning (CMI-MTL) model for Med-VQA, which consists of three key modules: the fine-grained visual-text feature alignment (FVTA) module, the cross-modal interleaved feature representation (CIFR) module, and the free-form answer enhanced multi-task learning (FFAE) module.
Motivated by Q-Former~\cite{li2023blip}, we introduce an FVTA module to effectively align the extracted visual and textual features through contrastive learning.
The FVTA generates learnable queries for better question-aware image representations. Additionally, it improves reasoning over complex medical questions, leading to more precise feature representations.
Inspired by the state-space model (SSM)~\cite{gu2023mamba}, a powerful successor to Transformers for sequential causal inference, we propose a CIFR module that dynamically captures cross-modal interactions between medical images and textual questions, producing coherent and context-aware feature representations. Consequently, CIFR can interpret complex medical questions that require sequential reasoning over global image features.
To fully utilize free-form open-ended questions, an FFAE module serves as an auxiliary component that generates answers for open-ended questions, complementing the mainstream classification branch within the multi-task learning framework. The FFAE further enhances the model's capability to handle open-ended Med-VQA.

The cross-modal data fusion method for images and text proposed in this paper can also spark further discussions and explorations beyond the medical-related fields, especially in areas such as model efficiency and cross-modal learning techniques. For example, graphics and image science focuses on understanding and modeling image scenes, and the cross-modal fusion of images and text can provide additional semantic information, thereby enhancing scene understanding~\cite{3DStyleGLIP}~\cite{Text2Mat}~\cite{ModelingSketches}. By combining visual data with the semantic context in text, we can more accurately identify objects and their relationships and facilitate an in-depth understanding of complex scenes.

The main contributions of our approach are summarized as: 
(1) \textbf{Research problem to be solved}: Beyond the general limitations in Med-VQA, we specifically address a more specific problem of leveraging open-ended questions as auxiliary information to enhance overall classification.
(2) \textbf{Innovative technical design}: Technically, we propose the CMI-MTL framework for medical vision-and-language understanding, integrating cross-Mamba interaction within a multi-task learning architecture.
(3) \textbf{Effectiveness and explainability}: Experimental results on the SLAKE, VQA-RAD, and OVQA datasets demonstrate the effectiveness of our model. Additionally, we perform an interpretability analysis using Grad-CAM~\cite{selvaraju2017grad} to assess whether the model focuses on the image regions most relevant to the questions when generating answers.

\section{Method}
\subsection{Problem formulation}
Similar to the general VQA task, Med-VQA can be formulated as a $C$-class classification problem for both open-ended and closed-ended questions.
Given a visual-text dataset $\mathcal{D}=\left\{(v_i,q_i,a_i)\right\}^{N}_{i=1}$, where $N$ denotes the number of the training samples, each triplet ($v_i, q_i, a_i$) represents the $i$-th medical image-question-answer pair.
The purpose of Med-VQA is to learn the mapping function $\mathcal{F}$ that predicts the answer $\hat{a}_i$ given an image-question pair $(v_i, q_i)$. This challenge can be formulated as maximizing the log-likelihood of the correct answer:
\begin{equation}
\hat{a}_i=\arg\max_{a_i\in \mathcal{A}}\mathcal{F}\left(a_i|v_i,q_i,\theta\right),
\end{equation}
where $\theta$ denotes model parameters, and $\mathcal{A}$ represents the set of candidate answers in $\mathcal{D}$.
\subsection{Fine-grained visual-text feature alignment}
As shown in Fig.~\ref{fig:architecture}(a), FVTA consists of an image encoder, a text encoder, and a question-aware Q-former (QQ-Former). The image encoder leverages ViT~\cite{dosovitskiy2021imageworth16x16words} to extract visual features from the input images, while the text encoder employs RoBERTa~\cite{liu2019robertarobustlyoptimizedbert} to extract the latent text embeddings. 

\textbf{Question-aware Q-former.} Inspired by Q-Former~\cite{li2023blip}, we propose QQ-Former to enhance the model's ability to generate accurate answers by identifying the most relevant parts of the question in relation to the image context. The QQ-former consists of a self-attention layer, a cross-attention layer, and a feed-forward network.

Formally, let $\{(v_i, q_i)\}_{i=1}^{B}$ represent a batch of $B$ visual-text pairs. The QQ-Former aligns visual features $\boldsymbol{v}_i$ extracted from ViT with text embeddings $\boldsymbol{t}_i \in \mathbb{R}^{d}$ obtained from RoBERTa, where $d$ denotes the embedding dimension. $\boldsymbol{v}_i$ interacts with a set of $K$ learnable query embeddings $\boldsymbol{z}_{i}\in\mathbb{R}^{K\times d}$ through cross-attention layers. The query embeddings $\boldsymbol{z}_{i}$ first pass through a self-attention layer before interacting with the visual features via a cross-attention layer. This process ensures that the query outputs, which integrate visual information, are aligned with the text representation $\boldsymbol{t}_i$.

To align the visual representation and text representation, we employ cross-modal contrastive learning (CMCL) to minimize the semantic similarity between images and texts.
 We denote $\boldsymbol{z}_{ij}\in{\mathbb{R}^{d}}$ as the $j$-th query token of the visual features $\boldsymbol{v}_i$. Visual-text similarity is measured by the maximum similarity between $\boldsymbol{t}$ and each row of $\boldsymbol{z}$. The final visual-text contrastive learning loss $\mathcal{L}_{\mathrm{vtc}}$ is the sum of the visual-to-text loss $\mathcal{L}_{\mathrm{v2t}}$ and the text-to-visual loss $\mathcal{L}_{\mathrm{t2v}}$, which can be written as:
\begin{equation}
\scriptsize
\begin{gathered}
    \mathcal{L}_{\mathrm{v2t}}=-\sum_{i=1}^K\log\frac{\exp(\cos(\boldsymbol{z}_{ij},\boldsymbol{t}_i)/\tau)}{\sum_{j=1}^B\exp(\cos(\boldsymbol{z}_{ij},\boldsymbol{t}_j)/\tau)}, 
    \\
    \mathcal{L}_{\mathrm{t2v}}=-\sum_{i=1}^{B}\log\frac{\exp(\cos(\boldsymbol{z}_{ji},\boldsymbol{t}_{i})/\tau)}{\sum_{j=1}^{K}\exp(\cos(\boldsymbol{z}_{ji},\boldsymbol{t}_i)/\tau)},
    \end{gathered}
\end{equation}
where $\cos(\cdot,\cdot)$ is the cosine similarity and $\tau$ is the temperature parameter.
\subsection{Cross-modal interleaved feature representation}
While Mamba-based methods demonstrate significant computational advantages compared to Transformer-based multimodal fusion methods, the Mamba's sequential scanning mechanism makes it challenging to effectively learn cross-modal correspondences. Thus, we propose cross-mamba interaction blocks in the CIFR module as shown in Fig.~\ref{fig:architecture}(b), aiming at exploiting interactions between learned visual representations and text embeddings while simultaneously capturing sequential relationships in the questions and learnable queries in the state-space.

\textbf{Cross-modal Mamba block.} 
Our approach proposes cross-modal Mamba (CMM) blocks, which preserve the sequential nature of Mamba and linear computational complexity while enabling effective cross-modal interactions. Given the normalized learnable queries $\boldsymbol{z}^{'}$ and the normalized text embeddings $\boldsymbol{t}^{'}$, we construct a pair of multimodal feature sequences, ($\boldsymbol{z}^{\text{t2z}}$, $\boldsymbol{t}^{\text{z2t}}$), by interleaving features from different modalities:
\begin{equation}
\footnotesize
\begin{gathered}
\boldsymbol{z}^{\text{t2z}}=\text{Fus}(\text{Mamba}(\boldsymbol{z}^{'}, \boldsymbol{z}^{'} \otimes \boldsymbol{t}^{'})) \oplus \boldsymbol{z}^{'},
\\
\boldsymbol{t}^{\text{z2t}}=\text{Fus}(\text{Mamba}(\boldsymbol{t}^{'}, \boldsymbol{t}^{'} \otimes \boldsymbol{z}^{'})) \oplus \boldsymbol{t}^{'},
\end{gathered}
\end{equation}
where $\text{Mamba}(\cdot,\cdot)$ denotes the original Mamba function, $\text{Fus}(\cdot,\cdot)$ represents the fusion function (implemented as a linear transformation in our work), $\otimes$ and $\oplus$ represent element-wise multiplication and element-wise summation, respectively.

The cross-mamba interaction representations for downstream tasks are obtained after two CMM blocks. Finally, we concatenate these features after two CMM blocks to obtain high-representative fused features, denoted as $\boldsymbol{x}^{\text{f}}$. The CMM enables a selective scan mechanism of Mamba to capture both intra- and inter-modal dependencies.

\begin{table*}[htb]
\centering
\caption{Comparisons with the state-of-the-art methods on three datasets. ${*}$ indicates that the data is copied from the original papers. The best values in \textbf{bold}. The second-best values are \underline{underlined}.}\label{tab:comparison}
    \begin{tabular}{c|ccc|ccc|ccc}
    \hline
        \multirow{2}{*}{Method}&   \multicolumn{3}{c|}{SLAKE}&\multicolumn{3}{c|}{VQA-RAD}&\multicolumn{3}{c}{OVQA}\\ \cline{2-10}

        &  Open&Closed&Overall&Open&Closed&  Overall&Open&Closed&  Overall \\
    \hline
    MEVF$^{*}$~\cite{nguyen2019overcoming}&-&-&-&37.41 &75.60 &60.42 &34.69&	74.21&58.46 \\
    
    MMQ$^{*}$~\cite{do2021multiple}&-&-&-&52.00	&76.71&	66.92&46.04&	76.13&64.14 \\
    
    M2I2~\cite{li2023self}&73.21&89.50&	77.60&48.78&	82.71&69.16&36.02&	72.38&57.89 \\
    MQAT~\cite{liu2022transformer}&77.20&87.97	&81.50&48.60&77.11&	65.50&38.21&79.92&67.00 \\
    
    VQAMix$^{*}$~\cite{gong2022vqamix}&-&-&-&56.90 &	79.50 &	70.50 &44.98&76.83&64.14\\
    MUMC~\cite{li2023masked}&75.04&93.27&82.19&62.57&84.19&75.61&39.05&75.87&61.20 \\
    
    MPR~\cite{ossowski2023retrieving}&75.50 &81.50 &77.90 &59.30 &79.10 &73.30 &31.70 &72.60 &56.97 \\
    
    LaPA~\cite{gu2024lapa}&80.00	&87.25	&82.84	&\underline{67.87}&	84.92& \textbf{78.15} &56.46 & 83.12 & 72.51\\
    
    M3AE~\cite{chen2024mapping}&\underline{80.78} &87.65 & \underline{83.47} &63.50 &85.42 &76.72 &\underline{61.65}&84.37& \underline{75.32} \\
    \hline
    CMI-MTL & \textbf{82.17}\(_{+1.59\uparrow}\) & 86.77 & \textbf{83.97} & \textbf{69.54}  & 81.37 & \underline{77.13} & \textbf{71.25} & 83.07 & \textbf{78.52} \\
    \hline
    \end{tabular}
\end{table*}

\subsection{Free-form answer enhanced multi-task learning}
We hypothesize that generating answers for open-ended questions is significantly more demanding than simply predicting a `yes' or `no' response. Strictly penalizing the model to encourage more precise free-form answer generation could enhance overall learning.

To this end, we design the FFAE module (Fig.~\ref{fig:architecture}(c)), which introduces an auxiliary branch to fully leverage knowledge from open-ended sequences, often overlooked by other methods. FFAE consists of two prediction heads: a classical classification head for both open-ended and closed-ended questions, and an auxiliary head dedicated to generating answers specifically for open-ended questions.

\textbf{Classification head.} The Classification head consists of a set of linear layers, a normalization layer,  and a GELU layer. The fused feature vector, $\boldsymbol{x}^{\text{f}}$, is classified using a binary cross-entropy loss function, denoted as $\mathcal{L}_{\text{cls}}$.

\textbf{Auxiliary head.} We adopt the pre-trained T5 decoder~\cite{raffel2020exploring} as the auxiliary head (AHead). We first filter out all the close-ended embeddings in the batch, obtaining the pure open-ended question embeddings $\{\boldsymbol{x}_{i}^{\text{f\_open}}\}_{i=1}^{O}$ with corresponding answers $\{a_i\}_{i=1}^{O}$, where $O$ is the number of open-ended samples in the batch ($O \leq B$). The T5 decoder generates corresponding answer prediction $\hat{a}_i$ for $\boldsymbol{x}_{i}^{\text{f\_open}}$. 
Next, we introduce a learnable attention mask $\mathcal{M}$, with all elements initially set to 1. This mask guides the model to focus on relevant tokens within the multi-modal features of open-ended questions.
Finally, we use mask-guided cross-entropy loss $\mathcal{L}_{\text{ce}}$ as our objective function:
\begin{equation}
    \mathcal{L}_{\text{aux}} = \mathcal{L}_{\text{ce}}(\hat{a}_i, \mathcal{M}, a_i).
\end{equation}
The multi-task learning loss is the summation of $\mathcal{L}_{\text{cls}}$ and $\mathcal{L}_{\text{aux}}$.

To this end, the overall loss function of our CMI-MTL is calculated by weighting $\mathcal{L}_{\text{cls}}$, $\mathcal{L}_{\text{vtc}}$, and $\mathcal{L}_{\text{aux}}$ as $\mathcal{L} = \mathcal{L}_{\text{cls}} + \alpha\mathcal{L}_{\text{vtc}} + \beta\mathcal{L}_{\text{aux}}$, where $\alpha$ and $\beta$ are the weight factors.

\section{Experiments}
\subsection{Datasets and setting}
\textbf{Dataset and metrics.} The \textbf{SLAKE}~\cite{liu2021slake} dataset supports both Chinese and English languages and covers a broad range of organ types. In this study, we focus on the English subset. SLAKE is partitioned at the image level, with 70\% (450 images) for training, 15\% (96 images) for validation, and 15\% (96 images) for testing following the approach in~\cite{chen2024mapping}. The \textbf{VQA-RAD}~\cite{lau2018dataset} dataset consists of radiological images annotated by volunteers with professional medical expertise. It includes 315 radiological images from three organ types and 3,515 clinical questions. To ensure fair comparisons with previous studies, we follow the established partitioning~\cite{li2023masked}, allocating 3,064 questions for training and 451 for testing. The \textbf{OVQA}~\cite{huang2022ovqa} dataset specializes in orthopedics, containing 19,020 medical visual question-answer pairs derived from 2,001 medical images. 
Following the original paper's protocol~\cite{van2023open}, we split OVQA into training (80\%, 15,216 questions), validation (10\%, 1,902 questions), and testing (10\%, 1,902 questions). The performance is evaluated using the Accuracy (ACC) metric.

The datasets used in this study, including the VQA-RAD dataset, encompass a range of clinically recognized rare or difficult diseases, such as posterior reversible encephalopathy syndrome (PRES), right aortic arch, pineal region mass, rare lesions in the cerebellopontine angle, and adult intussusception. These conditions, characterized by low incidence and complex clinical manifestations, represent the "rare or difficult cases" commonly encountered in clinical diagnosis. All these rare diseases have been incorporated into the model’s testing processes. The model’s high accuracy on the overall test set reflects its performance on these rare diseases. Moreover, our model demonstrates strong generalization ability, as evidenced by its extension to three widely used datasets for evaluating medical visual question answering tasks. The performances across all datasets further highlight its robust generalization capability. 
\begin{table*}[ht]
\centering
\caption{Performances of ablation study on multiple datasets.}\label{tab:ablation}
\begin{tabular}{cccc|ccc|ccc|ccc}
\hline
\multirow{2}{*}{QQ-Former} & \multirow{2}{*}{CMCL} & \multirow{2}{*}{CMM} & \multirow{2}{*}{AHead} & \multicolumn{3}{c|}{SLAKE} & \multicolumn{3}{c|}{VQA-RAD} & \multicolumn{3}{c}{OVQA} \\ \cline{5-13}
    & & & & Open & Closed & Overall & Open & Closed & Overall & Open & Closed & Overall \\
    
\hline
    &  &  &  & 80.31 & 85.25 & 82.25 & 60.52 & 	78.31 & 71.25 & 61.65 & 84.37 & 75.32  \\

    $\checkmark$ &  &  &  & 80.45 & 86.20 & 82.58 & 65.34 & 79.78 & 74.11 & 68.95 & 83.56 & 77.74  \\
    
    & $\checkmark$ &  &  & 80.72 & 85.96 & 82.77 & 64.77 & 80.51 & 74.33 & 66.09 & 84.42 & 77.07 \\

    &  & $\checkmark$ &  & 81.49 &85.96 &83.24 &65.34 &79.41 &73.88 & 68.07 & 83.74 & 77.48 \\

    &  &  & $\checkmark$ & 81.34 & 84.02 & 82.39 & 66.47 & 77.94 & 73.43 & 68.73 & 82.86 & 77.22 \\

    
    
    $\checkmark$ & $\checkmark$ & $\checkmark$ & $\checkmark$ & 82.17 & 86.77 & 83.97 & 69.54 & 81.37 & 77.13 & 71.25 & 83.07 & 78.52 \\
\hline
\end{tabular}
\end{table*}
\begin{figure*}[h]
    \centering
    \includegraphics[width=0.85\linewidth]{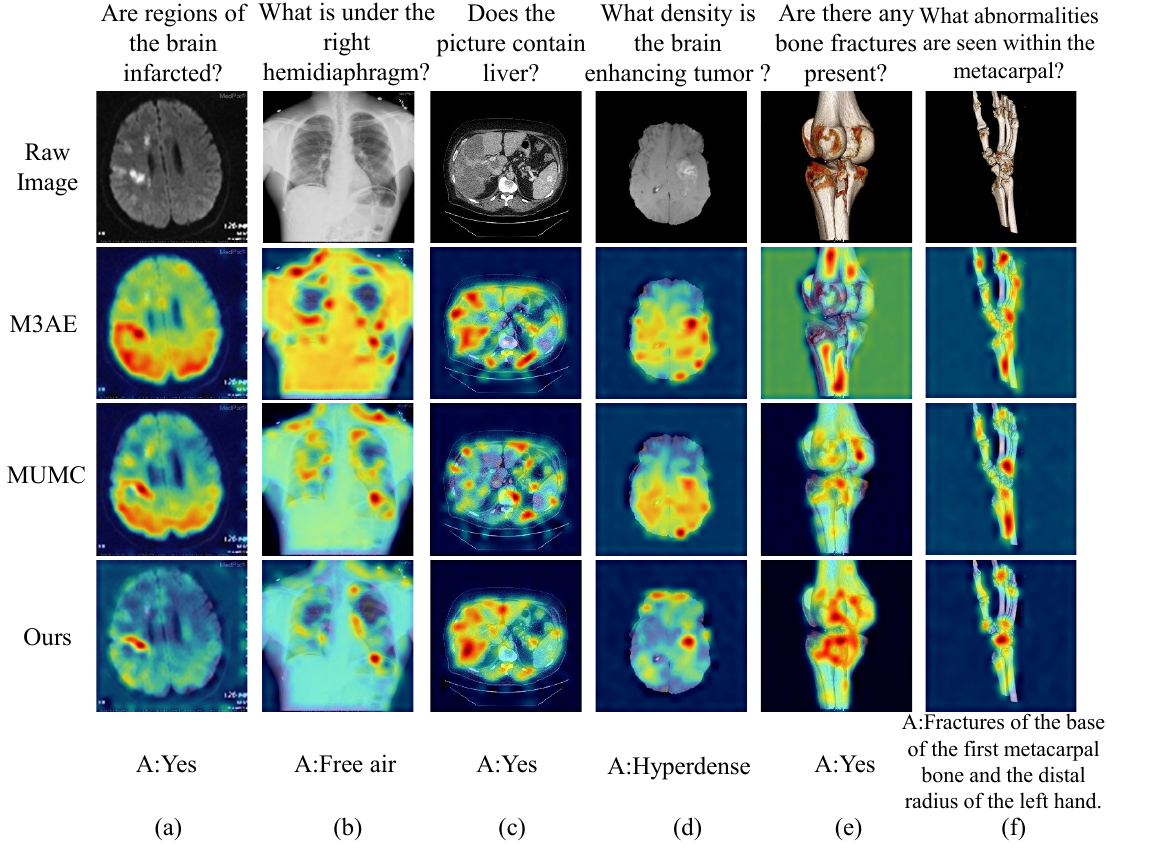}
    \caption{Visual saliency maps for representative methods.}
    \label{fig:visualizations}
\end{figure*}

\noindent{\textbf{Implementation details.}} Our method is implemented in PyTorch and executed on a single NVIDIA GeForce RTX 4090 GPU. The model is optimized using the AdamW optimizer with an initial learning rate of $5 \times 10^{-6}$. Training is conducted for 50 epochs with a batch size of 8. The hyperparameters $\alpha$ and $\beta$ are empirically set to 0.2 and 0.3, respectively. Medical images are resized to 384 $\times$ 384. The dimension of the learnable query vector is set to 32.

\subsection{Experimental Results}
\textbf{Comparison with the state-of-the-arts.} We re-implement several existing state-of-the-art methods, including pretraining and downstream task-based methods (M2I2~\cite{li2023self}, MUMC~\cite{li2023masked}, and M3AE~\cite{chen2024mapping}), knowledge-enhanced methods utilizing external data (MEVF~\cite{nguyen2019overcoming}, MPR~\cite{ossowski2023retrieving}, and LaPA~\cite{gu2024lapa}), a multiple metamodel quantification-based method (MMQ~\cite{do2021multiple}), a transformer-based method (MQAT~\cite{liu2022transformer}), and a data augmentation method without external data (VQAMix~\cite{gong2022vqamix}) to demonstrate the effectiveness of CMI-MTL. For models without publicly available source code, we directly report the results from their original papers.

Table~\ref{tab:comparison} presents a comparative analysis of our proposed method against state-of-the-art approaches on the SLAKE, VQA-RAD, and OVQA datasets. Specifically, on SLAKE, CMI-MTL achieves an overall ACC of 83.97\%, surpassing the second-best method, M3AE, which attains 83.47\%. On VQA-RAD, CMI-MTL achieves an overall ACC of 77.13\%, surpassing M3AE's 76.72\% and closely approaching LaPA's performance, which leads by only 1\%.

\begin{table*}[htb]
    \centering
    \caption{Comparisons of different $\alpha$ and $\beta$ values on the VQA-RAD dataset.}\label{tab:ablation_both}
        \begin{tabular}{c|ccc||c|ccc}
            \hline
            $\alpha$ & Open & Closed & Overall & $\beta$ & Open & Closed & Overall \\
            \hline
            0.1 & \textbf{69.89} & 80.15 & \underline{76.12} & 0.1 & 67.05 & 77.21 & 73.21 \\
            \textbf{0.2} & \underline{69.54} & \textbf{81.37} & \textbf{77.13} & 0.2 & \textbf{69.89} & \underline{79.41} & \underline{75.67} \\
            0.3 & \textbf{69.89} & \underline{79.41} & 75.67 & \textbf{0.3} & \underline{69.54} & \textbf{81.37} & \textbf{77.13} \\
            0.4 & 68.18 & 78.31 & 74.33 & 0.4 & \textbf{69.89} & 79.04 & 75.45 \\
            0.5 & 65.34 & 78.31 & 73.21 & 0.5 & 68.18 & 78.31 & 74.33 \\
            \hline
        \end{tabular}
\end{table*}

Notably, while LaPA, which makes use of the prior knowledge about the relationships between organs and diseases, excels in overall ACC, it does not achieve better performance than CMI-MTL in open-ended question answering. Similarly, on OVQA, CMI-MTL achieves an overall ACC of 78.52\%, outperforming M3AE's 75.32\% and LaPA's 72.51\%.

Notably, CMI-MTL achieves the best performance in open-ended question answering across all datasets. This can be attributed to the auxiliary FFAE module, which enhances the model's ability to handle open-ended medical VQA and improves its robustness in addressing diverse medical questions.
Moreover, CMI-MTL demonstrates superiority in both closed-ended and open-ended settings, reflecting the advantages of the coordinated interaction among FVTA, CIFR, and FFAE.

\textbf{Ablation study.} We conduct ablation studies to validate each proposed component on the SLAKE, VQA-RAD, and OVQA datasets by progressively adding modules, including QQ-Former, CMCL, CMM, and AHead, to the basic framework. As shown in Table~\ref{tab:ablation}, the basic method, enhanced with each module, achieves improvements across all datasets, particularly on VQA-RAD and OVQA. Specifically, the inclusion of the CMM module elevates ACC to 83.24\% on SLAKE, 73.88\% on VQA-RAD, and 77.48\% on OVQA, underscoring its effectiveness in capturing sequential relationships within image-text pairs. Furthermore, incorporating AHead in FFAE significantly improves accuracy in open-ended settings, yielding improvements of 1.03\%, 5.95\%, and 7.08\% over the basic model for SLAKE, VQA-RAD, and OVQA, respectively. These findings support our hypothesis that leveraging open-ended corpora can enhance learning in Med-VQA tasks.

\textbf{Effectiveness of weight factors.}
We study the effectiveness of weight factors in multi-task learning. As shown in Table~\ref{tab:ablation_both}, performance declines significantly when the weight factors $\alpha$ and $\beta$ exceed 0.2 and 0.3, respectively. This decline is likely due to the increased influence of the auxiliary head and contrastive loss affecting the main branch. When $\alpha$ and $\beta$ exceed 0.2 and 0.3, the positive impact of these auxiliary tasks becomes negative, disrupting the stable feature representation learned by the main branch. Based on these findings, we set the weight factors $\alpha$ and $\beta$ to 0.2 and 0.3 in the experiments.

\textbf{Interpretability analysis.}
We use Grad-CAM~\cite{selvaraju2017grad} to visualize the attention map between questions and images, analyzing the relevance of attended image regions, as shown in Fig.~\ref{fig:visualizations}. For closed-ended questions, CMI-MTL accurately focuses on the regions relevant to the questions, as illustrated in Fig.~\ref{fig:visualizations}(a), (c), and (e). For flexible open-ended questions, as shown in Fig.~\ref{fig:visualizations}(b) and (d), CMI-MTL effectively locates diseased regions in the image through reasoning based on the questions. For long-sequence answers, as shown in Fig.~\ref{fig:visualizations}(f), CMI-MTL precisely focuses on the subtle fractured area in the hand. It also includes challenging difficult cases. For instance, in the chest X-ray (Case b), artifacts cause blurring of features in the right subdiaphragmatic region. The attention heatmap of the model shows that its capture of the key feature "free air" is relatively scattered. Although the model attempts to focus on this feature, the response is weak due to the limitation of image quality. In cases of tiny cerebral infarcts (blurred-edged lesions in Case a) and atypical hand fractures (subtle bone cracks in Case f), the model's heatmap can roughly locate the regions but struggles to finely distinguish the "tiny features". Despite these challenges, our model still outperforms other state-of-the-art methods. These results demonstrate the strong reasoning ability and effectiveness of the proposed CMI-MTL model, as well as its capability to accurately identify regions of interest in the questions.

\textbf{Efficiency analysis.}
We evaluate the efficiency of Mamba and Transformer using three key metrics: parameter count, FLOPs, and memory consumption. Typically, compared to the Transformer architecture (using M3AE as a reference, with 345M parameters, 81.7G FLOPs, and 21 GB memory), our Mamba-based CMI-MTL demonstrates significant improvements in computational complexity, memory consumption, and operational speed, with 245M parameters, 74.2G FLOPs, and 19 GB memory. These improvements are primarily due to the core architectural advantage of Mamba, which eliminates redundant parameters in the "query/key/value" (QKV) projection layers of the multi-head attention module in Transformer. The efficiency comparison highlights that our model effectively balances performance and complexity, enabling efficient training and inference while maintaining stable accuracy. 

\section{Conclusion}
In conclusion, we introduce a novel Cross-Mamba Interaction based Multi-Task Learning framework for Med-VQA. Our model comprises fine-grained visual-text feature alignment, cross-modal interleaved feature representation, and free-form answer enhanced multi-task learning. These components facilitate effective interaction between image-text pairs and fully leverage auxiliary open-ended question data to enhance overall learning.
Comprehensive experiments and analyses demonstrate the superiority of the proposed method. Furthermore, interpretability analysis highlights the model's ability to focus on relevant image regions when answering questions. Future work will explore extending our approach to larger datasets in the Med-VQA domain.

Since the initial phase of this study focused on optimizing the model architecture and verifying model performance on public benchmarks, it did not include concurrent evaluation data from actual medical practitioners. In subsequent research, we will prioritize "human-machine comparison and collaboration" as a key focus. On one hand, we will expand the sample size to conduct multi-center validation, incorporate evaluation results from doctors in hospitals of different levels, and analyze the model's applicability in various clinical environments. On the other hand, we will explore human-machine collaborative diagnosis models, assess the role of model assistance in improving doctors' diagnostic performance, aiming to be more in line with real clinical application scenarios. 

\bibliographystyle{eg-alpha-doi} 
\bibliography{egbibsample}       

\newcommand{\etalchar}[1]{$^{#1}$}
\begin{thebibliography}{\uppercase{LGBADF18}}

\bibitem[ADL{\etalchar{*}}22]{alayrac2022flamingo}
\textsc{Alayrac J.-B., Donahue J., Luc P., Miech A., Barr I., Hasson Y., Lenc K., Mensch A., Millican K., Reynolds M., et~al.}:
\newblock Flamingo: a visual language model for few-shot learning.
\newblock \emph{Advances in Neural Information Processing Systems 35} (2022), 23716--23736.

\bibitem[BAHD{\etalchar{*}}19]{ben2019vqa}
\textsc{Ben~Abacha A., Hasan S.~A., Datla V.~V., Demner-Fushman D., M{\"u}ller H.}:
\newblock {VQA-M}ed: Overview of the medical visual question answering task at {I}mage{CLEF} 2019.
\newblock In \emph{Proceedings of CLEF (Conference and Labs of the Evaluation Forum) 2019 Working Notes} (2019), 9-12 September 2019.

\bibitem[BFH{\etalchar{*}}98]{Buhmann:1998:DCQ}
\textsc{Buhmann J.~M., Fellner D.~W., Held M., Ketterer J., Puzicha J.}:
\newblock Dithered color quantization.
\newblock \emph{Computer Graphics Forum 17}, 3 (Sept. 1998), C219--C231.
\newblock (Proc. Eurographics'98) \httpsAddr{//diglib.eg.org/handle/10.2312/8491}.
\newblock \href {https://doi.org/10.1111/1467-8659.00269} {\path{doi:10.1111/1467-8659.00269}}.

\bibitem[CBHX24]{chen2024multi}
\textsc{Chen Q., Bian M., He W., Xu H.}:
\newblock Multi-modal multi-scale state space model for medical visual question answering.
\newblock In \emph{International Conference on Artificial Neural Networks} (2024), Springer, pp.~328--342.

\bibitem[CDH{\etalchar{*}}24]{chen2024mapping}
\textsc{Chen Z., Du Y., Hu J., Liu Y., Li G., Wan X., Chang T.-H.}:
\newblock Mapping medical image-text to a joint space via masked modeling.
\newblock \emph{Medical Image Analysis 91} (2024), 103018.

\bibitem[CPK24]{3DStyleGLIP}
\textsc{Chung S., Park J., Kang H.}:
\newblock {3DStyleGLIP: Part-Tailored Text-Guided 3D Neural Stylization}.
\newblock In \emph{Pacific Graphics Conference Papers and Posters} (2024), Chen R., Ritschel T., Whiting E., (Eds.), The Eurographics Association.
\newblock \href {https://doi.org/10.2312/pg.20241320} {\path{doi:10.2312/pg.20241320}}.

\bibitem[DBK{\etalchar{*}}21]{dosovitskiy2021imageworth16x16words}
\textsc{Dosovitskiy A., Beyer L., Kolesnikov A., Weissenborn D., Zhai X., Unterthiner T., Dehghani M., Minderer M., Heigold G., Gelly S., Uszkoreit J., Houlsby N.}:
\newblock An image is worth 16x16 words: Transformers for image recognition at scale.
\newblock In \emph{ICLR} (2021).

\bibitem[DNT{\etalchar{*}}21]{do2021multiple}
\textsc{Do T., Nguyen B.~X., Tjiputra E., Tran M., Tran Q.~D., Nguyen A.}:
\newblock Multiple meta-model quantifying for medical visual question answering.
\newblock In \emph{International Conference on Medical Image Computing and Computer-Assisted Intervention.} (2021), Springer, pp.~64--74.

\bibitem[DZL{\etalchar{*}}24]{dong2024fusion}
\textsc{Dong W., Zhu H., Lin S., Luo X., Shen Y., Liu X., Zhang J., Guo G., Zhang B.}:
\newblock Fusion-mamba for cross-modality object detection.
\newblock \emph{arXiv preprint arXiv:2404.09146} (2024).

\bibitem[FH93]{Fellner-Helmberg93}
\textsc{Fellner D.~W., Helmberg C.}:
\newblock Robust rendering of general ellipses and elliptical arcs.
\newblock \emph{ACM TOG 12}, 3 (July 1993), 251--276.
\newblock \href {https://doi.org/10.1145/169711.169704} {\path{doi:10.1145/169711.169704}}.

\bibitem[FvDF{\etalchar{*}}93]{FolDamFeiHug.etal93}
\textsc{Foley J.~D., van Dam A., Feiner S.~K., Hughes J.~F., Phillips R.}:
\newblock \emph{Introduction to Computer Graphics}.
\newblock Addison-Wesley, 1993.

\bibitem[GCM{\etalchar{*}}22]{gong2022vqamix}
\textsc{Gong H., Chen G., Mao M., Li Z., Li G.}:
\newblock {VQAM}ix: Conditional triplet mixup for medical visual question answering.
\newblock \emph{IEEE Transactions on Medical Imaging 41}, 11 (2022), 3332--3343.

\bibitem[GD23]{gu2023mamba}
\textsc{Gu A., Dao T.}:
\newblock Mamba: Linear-time sequence modeling with selective state spaces.
\newblock \emph{arXiv preprint arXiv:2312.00752} (2023).

\bibitem[GYLC24]{gu2024lapa}
\textsc{Gu T., Yang K., Liu D., Cai W.}:
\newblock Lapa: Latent prompt assist model for medical visual question answering.
\newblock In \emph{Proceedings of the IEEE/CVF Conference on Computer Vision and Pattern Recognition} (2024), pp.~4971--4980.

\bibitem[HCZ{\etalchar{*}}25]{he2025pan}
\textsc{He X., Cao K., Zhang J., Yan K., Wang Y., Li R., Xie C., Hong D., Zhou M.}:
\newblock Pan-mamba: Effective pan-sharpening with state space model.
\newblock \emph{Information Fusion 115} (2025), 102779.

\bibitem[HGZ{\etalchar{*}}23]{Text2Mat}
\textsc{He Z., Guo J., Zhang Y., Tu Q., Chen M., Guo Y., Wang P., Dai W.}:
\newblock {Text2Mat: Generating Materials from Text}.
\newblock In \emph{Pacific Graphics Short Papers and Posters} (2023), Chaine R., Deng Z., Kim M.~H., (Eds.), The Eurographics Association.
\newblock \href {https://doi.org/10.2312/pg.20231275} {\path{doi:10.2312/pg.20231275}}.

\bibitem[HWLH22]{huang2022ovqa}
\textsc{Huang Y., Wang X., Liu F., Huang G.}:
\newblock {OVQA}: A clinically generated visual question answering dataset.
\newblock In \emph{Proceedings of the 45th International ACM SIGIR Conference on Research and Development in Information Retrieval} (2022), pp.~2924--2938.

\bibitem[JLL{\etalchar{*}}24]{ModelingSketches}
\textsc{Jing J., Liu Y., Li M., Xiao Q., Chai S.}:
\newblock {Modeling Sketches both Semantically and Structurally for Zero-Shot Sketch-Based Image Retrieval is Better}.
\newblock In \emph{Pacific Graphics Conference Papers and Posters} (2024), Chen R., Ritschel T., Whiting E., (Eds.), The Eurographics Association.
\newblock \href {https://doi.org/10.2312/pg.20241309} {\path{doi:10.2312/pg.20241309}}.

\bibitem[KBM{\etalchar{*}}21]{9434063}
\textsc{Khare Y., Bagal V., Mathew M., Devi A., Priyakumar U.~D., Jawahar C.}:
\newblock Mmbert: Multimodal bert pretraining for improved medical vqa.
\newblock In \emph{2021 IEEE 18th International Symposium on Biomedical Imaging (ISBI)} (2021), pp.~1033--1036.
\newblock \href {https://doi.org/10.1109/ISBI48211.2021.9434063} {\path{doi:10.1109/ISBI48211.2021.9434063}}.

\bibitem[KSS97]{Kobbelt97-USHDR}
\textsc{Kobbelt L., Stamminger M., Seidel H.-P.}:
\newblock Using subdivision on hierarchical data to reconstruct radiosity distribution.
\newblock \emph{Computer Graphics Forum 16}, 3 (1997), C347--C355.
\newblock (Proc. Eurographics'97) \httpsAddr{//diglib.eg.org/handle/10.2312/8393}.
\newblock \href {https://doi.org/10.1111/1467-8659.16.3conferenceissue.36} {\path{doi:10.1111/1467-8659.16.3conferenceissue.36}}.

\bibitem[LFTG97]{Lafortune97-NARF}
\textsc{Lafortune E.~P., Foo S.-C., Torrance K.~E., Greenberg D.~P.}:
\newblock Non-linear approximation of reflectance functions.
\newblock In \emph{Proc. SIGGRAPH '97} (1997), vol.~31, pp.~117--126.
\newblock \href {https://doi.org/10.1145/258734.258801} {\path{doi:10.1145/258734.258801}}.

\bibitem[LGBADF18]{lau2018dataset}
\textsc{Lau J.~J., Gayen S., Ben~Abacha A., Demner-Fushman D.}:
\newblock A dataset of clinically generated visual questions and answers about radiology images.
\newblock \emph{Scientific Data 5}, 1 (2018), 1--10.

\bibitem[LHZ{\etalchar{*}}24]{10256025}
\textsc{Liu J., Hu T., Zhang Y., Feng Y., Hao J., Lv J., Liu Z.}:
\newblock Parameter-efficient transfer learning for medical visual question answering.
\newblock \emph{IEEE Transactions on Emerging Topics in Computational Intelligence 8}, 4 (2024), 2816--2826.
\newblock \href {https://doi.org/10.1109/TETCI.2023.3311333} {\path{doi:10.1109/TETCI.2023.3311333}}.

\bibitem[LLH{\etalchar{*}}23]{li2023masked}
\textsc{Li P., Liu G., He J., Zhao Z., Zhong S.}:
\newblock Masked vision and language pre-training with unimodal and multimodal contrastive losses for medical visual question answering.
\newblock In \emph{International Conference on Medical Image Computing and Computer-Assisted Intervention} (2023), Springer, pp.~374--383.

\bibitem[LLSH23]{li2023blip}
\textsc{Li J., Li D., Savarese S., Hoi S.}:
\newblock Blip-2: Bootstrapping language-image pre-training with frozen image encoders and large language models.
\newblock In \emph{International Conference on Machine Learning} (2023), PMLR, pp.~19730--19742.

\bibitem[LLT{\etalchar{*}}23]{li2023self}
\textsc{Li P., Liu G., Tan L., Liao J., Zhong S.}:
\newblock Self-supervised vision-language pretraining for medial visual question answering.
\newblock In \emph{2023 IEEE 20th International Symposium on Biomedical Imaging (ISBI)} (2023), IEEE, pp.~1--5.

\bibitem[LOG{\etalchar{*}}19]{liu2019robertarobustlyoptimizedbert}
\textsc{Liu Y., Ott M., Goyal N., Du J., Joshi M., Chen D., Levy O., Lewis M., Zettlemoyer L., Stoyanov V.}:
\newblock {RoBERTa: A Robustly Optimized BERT Pretraining Approach}.
\newblock \emph{arXiv preprint arXisv:1907.11692} (2019).

\bibitem[Lou90]{Lous90}
\textsc{Lous Y.~L.}:
\newblock Report on the {F}irst {E}urographics {W}orkshop on {V}isualization in {S}cientific {C}omputing.
\newblock \emph{Computer Graphics Forum 9}, 4 (Dec. 1990), 371--372.
\newblock \href {https://doi.org/10.1111/j.1467-8659.1990.tb00430.x} {\path{doi:10.1111/j.1467-8659.1990.tb00430.x}}.

\bibitem[LPZ{\etalchar{*}}24]{li2024mambadfuse}
\textsc{Li Z., Pan H., Zhang K., Wang Y., Yu F.}:
\newblock Mambadfuse: A mamba-based dual-phase model for multi-modality image fusion.
\newblock \emph{arXiv preprint arXiv:2404.08406} (2024).

\bibitem[LSGZ22]{liu2022transformer}
\textsc{Liu L., Su X., Guo H., Zhu D.}:
\newblock A transformer-based medical visual question answering model.
\newblock In \emph{2022 26th International Conference on Pattern Recognition (ICPR)} (2022), IEEE, pp.~1712--1718.

\bibitem[LZT{\etalchar{*}}23]{lin2023medical}
\textsc{Lin Z., Zhang D., Tao Q., Shi D., Haffari G., Wu Q., He M., Ge Z.}:
\newblock Medical visual question answering: A survey.
\newblock \emph{Artificial Intelligence in Medicine 143} (2023), 102611.

\bibitem[LZX{\etalchar{*}}21]{liu2021slake}
\textsc{Liu B., Zhan L.-M., Xu L., Ma L., Yang Y., Wu X.-M.}:
\newblock {SLAKE}: A semantically-labeled knowledge-enhanced dataset for medical visual question answering.
\newblock In \emph{2021 IEEE 18th International Symposium on Biomedical Imaging (ISBI)} (2021), IEEE, pp.~1650--1654.

\bibitem[NDN{\etalchar{*}}19]{nguyen2019overcoming}
\textsc{Nguyen B.~D., Do T.-T., Nguyen B.~X., Do T., Tjiputra E., Tran Q.~D.}:
\newblock Overcoming data limitation in medical visual question answering.
\newblock In \emph{International Conference on Medical Image Computing and Computer-Assisted Intervention.} (2019), Springer, pp.~522--530.

\bibitem[OH23]{ossowski2023retrieving}
\textsc{Ossowski T., Hu J.}:
\newblock Retrieving multimodal prompts for generative visual question answering.
\newblock In \emph{Findings of the Association for Computational Linguistics: ACL 2023} (2023), pp.~2518--2535.

\bibitem[PHZ{\etalchar{*}}22]{PAN2022109763}
\textsc{Pan H., He S., Zhang K., Qu B., Chen C., Shi K.}:
\newblock Amam: An attention-based multimodal alignment model for medical visual question answering.
\newblock \emph{Knowledge-Based Systems 255} (2022), 109763.
\newblock \href {https://doi.org/https://doi.org/10.1016/j.knosys.2022.109763} {\path{doi:https://doi.org/10.1016/j.knosys.2022.109763}}.

\bibitem[PKR{\etalchar{*}}18]{pelka2018radiology}
\textsc{Pelka O., Koitka S., R{\"u}ckert J., Nensa F., Friedrich C.~M.}:
\newblock Radiology objects in context (roco): a multimodal image dataset.
\newblock In \emph{Intravascular Imaging and Computer Assisted Stenting and Large-Scale Annotation of Biomedical Data and Expert Label Synthesis: 7th Joint International Workshop, CVII-STENT 2018 and Third International Workshop, LABELS 2018, Held in Conjunction with MICCAI 2018, Granada, Spain, September 16, 2018, Proceedings 3} (2018), Springer, pp.~180--189.

\bibitem[RSR{\etalchar{*}}20]{raffel2020exploring}
\textsc{Raffel C., Shazeer N., Roberts A., Lee K., Narang S., Matena M., Zhou Y., Li W., Liu P.~J.}:
\newblock Exploring the limits of transfer learning with a unified text-to-text transformer.
\newblock \emph{Journal of Machine Learning Research 21}, 140 (2020), 1--67.

\bibitem[SCD{\etalchar{*}}17]{selvaraju2017grad}
\textsc{Selvaraju R.~R., Cogswell M., Das A., Vedantam R., Parikh D., Batra D.}:
\newblock Grad-cam: Visual explanations from deep networks via gradient-based localization.
\newblock In \emph{2017 IEEE International Conference on Computer Vision (ICCV)} (2017), pp.~618--626.

\bibitem[Sei93]{Seidel93}
\textsc{Seidel H.-P.}:
\newblock Polar forms for geometrically continuous spline curves of arbitrary degree.
\newblock \emph{ACM TOG 12}, 1 (Jan. 1993), 1--34.
\newblock \href {https://doi.org/10.1145/169728.169726} {\path{doi:10.1145/169728.169726}}.

\bibitem[SIKR22]{seenivasan2022surgical}
\textsc{Seenivasan L., Islam M., Krishna A.~K., Ren H.}:
\newblock Surgical-vqa: Visual question answering in surgical scenes using transformer.
\newblock In \emph{International Conference on Medical Image Computing and Computer-Assisted Intervention} (2022), Springer, pp.~33--43.

\bibitem[VSDN{\etalchar{*}}23]{van2023open}
\textsc{Van~Sonsbeek T., Derakhshani M.~M., Najdenkoska I., Snoek C.~G., Worring M.}:
\newblock Open-ended medical visual question answering through prefix tuning of language models.
\newblock In \emph{International Conference on Medical Image Computing and Computer-Assisted Intervention} (2023), Springer, pp.~726--736.

\bibitem[VSP{\etalchar{*}}17]{vaswani2017attention}
\textsc{Vaswani A., Shazeer N., Parmar N., Uszkoreit J., Jones L., Gomez A.~N., Kaiser L., Polosukhin I.}:
\newblock Attention is all you need.
\newblock In \emph{Proceedings of the 31st International Conference on Neural Information Processing Systems} (2017), p.~6000–6010.

\bibitem[ZZC{\etalchar{*}}24]{zhao2024topicwise}
\textsc{Zhao J., Zhou Y., Chen Z., Fu H., Wan L.}:
\newblock Topicwise separable sentence retrieval for medical report generation.
\newblock \emph{arXiv preprint arXiv:2405.04175} (2024).

\end{thebibliography}


\end{document}